\documentclass[sigconf]{acmart}
\renewcommand\footnotetextcopyrightpermission[1]{} % removes footnote with conference information in first column
\AtBeginDocument{%
  \providecommand\BibTeX{{%
    \normalfont B\kern-0.5em{\scshape i\kern-0.25em b}\kern-0.8em\TeX}}}

\usepackage{algorithm}
\usepackage{amsmath}
\usepackage{algpseudocode}
\usepackage{booktabs}

\usepackage{amssymb}
\usepackage{color}
\usepackage{amsfonts,amssymb}
\begin{document}

\title{Trusted Multi-Scale Classification Framework \\ for Whole Slide Image}

\author{Ming Feng}
\affiliation{%
  %\department{College of Electronic and Information Engineering}
  \institution{Tongji University}
  % \city{Shanghai}
  \country{}
  }
\email{1810865@tongji.edu.cn}

\author{Kele Xu}
\authornote{Corresponding author.}
\affiliation{%
  \institution{National University of Defense Technology}
  % \city{Changsha}
  \country{}
  }
\email{xukelele@163.com}

\author{Nanhui Wu}
\affiliation{%
  %\department{College of Electronic and Information Engineering}
  \institution{Shanghai Skin Disease Hospital}
  % \city{Shanghai}
  \country{}
  }
\email{summer\_xiasen@126.com}

\author{Weiquan Huang}
\affiliation{%
  %\department{College of Electronic and Information Engineering}
  \institution{Tongji University}
  \country{}
  }
\email{weiquanh@tongji.edu.cn}

\author{Yan Bai}
\affiliation{%
  %\department{College of Electronic and Information Engineering}
  \institution{Tongji University}
  % \city{Shanghai}
  \country{}
  }
\email{yan.bai@tongji.edu.cn}

\author{Changjian Wang}
\affiliation{%
  \institution{National University of Defense Technology}
  % \city{Changsha}
  \country{}
  }
\email{andrew_tal@yeah.net}

\author{Huaimin Wang}
\affiliation{%
  \institution{National University of Defense Technology}
  % \city{Changsha}
  \country{}
  }
\email{whm_w@163.com}


\begin{abstract}
Despite remarkable efforts been made, the classification of gigapixels whole-slide image (WSI) is severely restrained from either the constrained computing resources for the whole slides, or limited utilizing of the knowledge from different scales. Moreover, most of the previous attempts lacked of the ability of uncertainty estimation.
Generally, the pathologists often jointly analyze WSI from the different magnifications. If the pathologists are uncertain by using single magnification, then they will change the magnification repeatedly to discover various features of the tissues. Motivated by the diagnose process of the pathologists, in this paper, we propose a trusted multi-scale classification framework for the WSI. Leveraging the Vision Transformer as the backbone for multi branches, our framework can jointly classification modeling, estimating the uncertainty of each magnification of a microscope and integrate the evidence from different magnification.
Moreover, to exploit discriminative patches from WSIs and reduce the requirement for computation resources, we propose a novel patch selection schema using attention rollout and non-maximum suppression.
To empirically investigate the effectiveness of our approach, empirical experiments are conducted on our WSI classification tasks, using two benchmark databases. The obtained results suggest that the trusted framework can significantly improve the WSI classification performance compared with the state-of-the-art methods.
\end{abstract}

\begin{CCSXML}
<ccs2012>
   <concept>
       <concept_id>10010405.10010444.10010447</concept_id>
       <concept_desc>Applied computing~Health care information systems</concept_desc>
       <concept_significance>500</concept_significance>
       </concept>
   <concept>
       <concept_id>10002951.10003227.10003351</concept_id>
       <concept_desc>Information systems~Data mining</concept_desc>
       <concept_significance>500</concept_significance>
       </concept>
 </ccs2012>
\end{CCSXML}

\ccsdesc[500]{Applied computing~Health care information systems}
\ccsdesc[500]{Information systems~Data mining}

\keywords{Histopathological Images Analysis, Whole Slide Images, Trusted Multi-Scale Classification}


\maketitle

\section{Introduction}
% Problem definition
The interpretation of pathological images has drawn increasing attentions during last decades, as it is essential to understand the disease progression \cite{campanella2019clinical} and assist the clinicians in making early decision on treatments \cite{wang2019weakly}. 
In the practical settings, the tissue slides are routinely examined by pathologists manually which is time-consuming. Thus, it is desirable to develop an automated analysis method, which could considerably speed up the diagnosis and facilitate the in-time treatment \cite{wang2019weakly}.

% Running example

\begin{figure}[!t]
	\centering
	\includegraphics[scale=0.45]{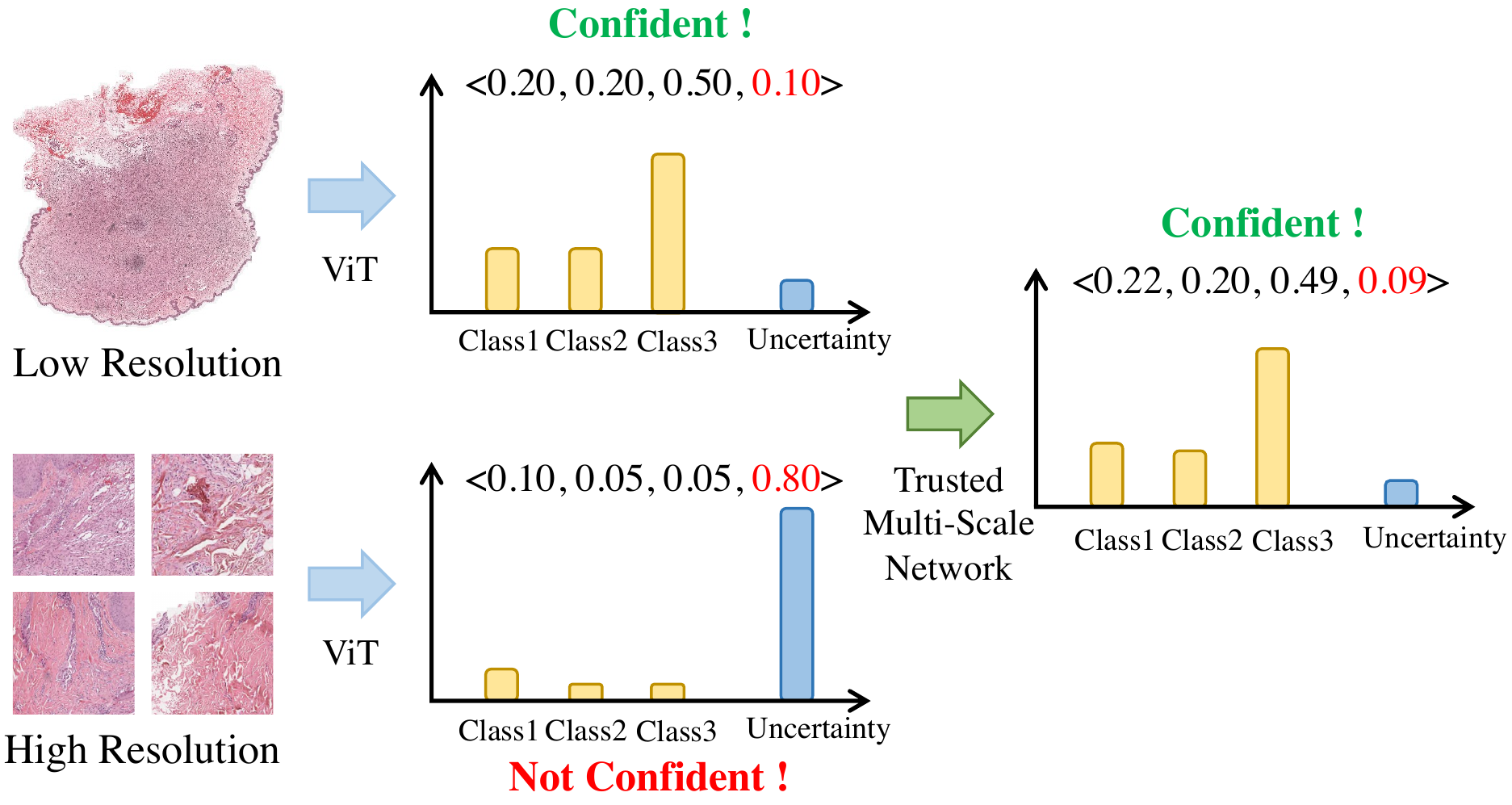}
	\caption{The decision-making process of trusted multi-scale classification. At high resolution, we hope to reduce the influence on the final decision when the discriminative patches are not selected.}
	\label{fig.aim}
\end{figure}

% Previous attempts
Last decade, convolutional neural networks (CNNs) have dominated the fields of histopathology image interpretation \cite{araujo2017classification} and sustainable efforts have been made to improve the different tasks' performance using different network architectures. Presently, many efforts focus on patch-based methods \cite{hashimoto2020multi}, combined with multi instance-learning (MIL) for the WSI classification task. It is evident that high resolution patch-level analysis can be helpful to improve the performance, however, it is computation intensive and can substantially increase processing time \cite{maksoud2020sos}. To reduce the demand of the computation resources, \cite{maksoud2020sos} proposed a selection method to find out the discriminative patches.
Despite the efforts have been made, the analysis of pathological images is still confronted with several challenges:
\begin{itemize}
	\item Firstly, the pathologists will often identify the tumor region and perform subtype classification by carefully considering the information from different magnifications. If the pathologist is uncertain using one magnification, then they will repeatedly change the magnification of a microscope to discover the features from multi-scales. Most of current WSI framework cannot mimic these pathologist’s actual practices as these WSI frameworks are lack of the ability for uncertainty estimation.
	
	\item Secondly, as aforementioned, conventional approaches cannot process the gigapixel resolution of the WSI due to the constraint of computation resource. Selecting the representative patch is a good solution candidate. However, the selecting standard has a great influence on the classification performance. How to select the discriminative patch is still under explored in previous studies.
	
	\item Thirdly, the modeling in pathological images analysis has been dominated by CNNs. Recently, the Transformer architecture \cite{liu2021swin} has demonstrated promising results on certain tasks, including the image classification and joint vision-language modeling task. Few attempts have been made to expand the applicability of Transformer such that it can serve as backbone architectures for WSI analysis.
\end{itemize}

To address aforementioned challenges and develop a practical WSI classification system, this paper proposes a novel framework from the trusted perspective using the Vision Transformer architecture. Specifically, the contributions of this paper are summarized as follows:

\begin{itemize}
	\item Firstly, to mimic the pathologist’s actual practices and the automatic WSI classification system, we propose a multi-scale WSI classification framework from the trusted perspective. Leveraging the vision Transformer as the backbone for different branches, our framework can jointly classification modeling and estimating the uncertainty of each magnification of a microscope and integrate the evidence from different magnification (as shown in Fig. \ref{fig.aim}).
	
	% TBD 
	\item Secondly, we introduce a novel patch selection procedure using attention rollout, to eliminate these redundant patches. Compared to the-state-of-art selection method, our method can provide higher robustness and effectiveness, while requires less computation time.
	
	\item Thirdly, to validate our method, empirical experiments have been conducted on the WSI classification task, using two databases: Liver-Kidney-Stomach immunofluorescence WSIs \cite{maksoud2020sos} and Fibroma hematoxylin-eosin WSIs. The obtained results demonstrate that our method can improve state-of-the-art by a big margin. 
\end{itemize}

The following of this paper is organized as follows. Section 2 discusses the relationship between our method and previous work, while, our methodology is described in section 3. The comprehensive experimental results are given in section 4 and Section 5 draws the conclusion.

\section{Related Works}
\subsection{Whole-Slide Image Analysis}
Since the revolution of neural network \cite{lecun2015deep}, CNNs have became dominated this field and been applied for different tasks \cite{srinidhi2020deep,hanna2019whole}. However, it is widely known that directly apply a CNN on the WSI (usually on the gigabyte resolution) is currently computationally impossible, due to the severely constraint of the hardware. In \cite{hou2016patch}, the authors proposed a two-stages model for WSI classification which employs a patch-level CNN and builds a decision fusion model in the second stage. Many literates find that: the weakly supervised learning \cite{wang2019weakly,zhu2017wsisa} can be used to improve the WSI classification performance. To reduce the demand of the computation resources, the authors in \cite{maksoud2020sos} proposed a selection method to find out the discriminative patches. The modeling in pathological images analysis has been dominated by CNNs while few attempts have been made to expand the applicability of Transformer \cite{dosovitskiy2020image,liu2021swin} for WSI analysis. 

\subsection{Multi-Scale Methods}
Single patch can provide limited contextual information of the WSIs while processing the WSIs in one step is computationally impossible. 
Multi-scale networks can be used to resolve this problem by using multiple inputs from different scales, with the goal to capture different levels of detail in the WSIs. \cite{ghafoorian2017deep} proposed a later fusion method for the WSIs. Conventional multi-scales contains large redundancies which can increase the computation demand. In \cite{dong2018reinforced}, the authors explored the reinforced learning, to determine whether to use high or low magnifications to process the WSIs. Many of the traditional multi-scales methods cannot copy with the redundancies efficiently, thus require extra computational resources.
The authors in \cite{maksoud2020sos} proposed a dynamic multi-scale methods, named SOS, to eliminate the redundancies.

% TBD (CCA)
\subsection{Trusted Multi-View Classification}
Multi-View classification itself is a well-established topic \cite{zhang2020deep,kan2016multi}, we refer the reader to find more details in the extensive survey \cite{carbonneau2018multiple}. To handle the discrepancy between different views, one natural way is to project all views into a common space, for example, the Canonical Correlation Analysis (CCA). The authors in \cite{yang2019adaptive} proposed an adaptive-weighing mechanism for the multi-view classification while \cite{zhang2017hierarchical} explored a novel representation learning method for the classification task from the implicitly perspective. However, most of the previous attempts for multi-view classifications neglect the fact that: the quality of a view for different samples can be different. Thus, the models cannot provide reliable uncertainty estimations which indicate whether predictions can be trusted.
To address this challenge, the authors in \cite{han2021trusted} a trusted multi-view learning method by dynamically integrating different views at an evidence level. Very few attempts explore to extend the multi-view classification for the WSIs, as the aggregation between different scales can be challenging.

\begin{figure*}[!htbp]
	\centering
	\includegraphics[scale=0.55]{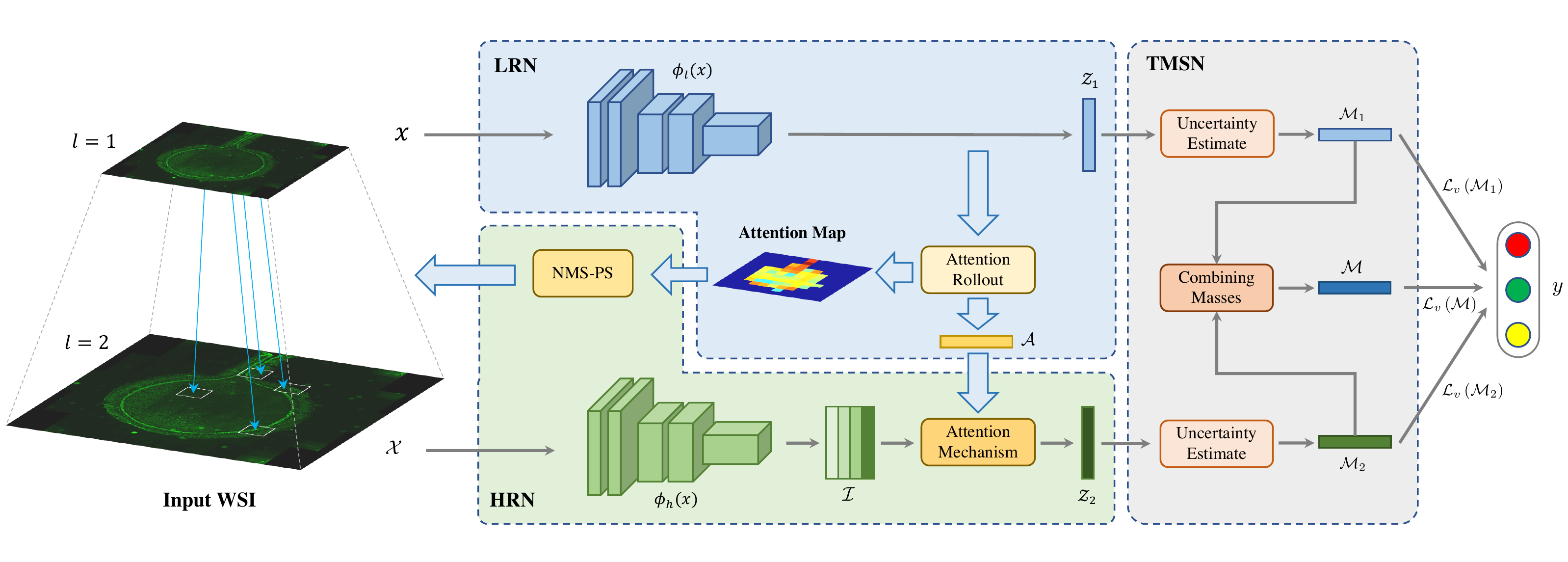}
	\caption{The framework of our purposed model. $x$ represents the thumbnail of WSI. Firstly, we input $x$ into \textbf{LRN} (Low Resolution Network, denoted as $\phi_{l}$) to obtain low-level features $z_{1}$. Then, under the guidance of \textbf{AR} (attention rollout) and \textbf{NMS-PS} (Non-Maximum Suppression Patch Selection), attentive patches $\mathcal{X}$ can be extracted from high resolution, followed by \textbf{HRN} (High Resolution Network, denoted as $\phi_{h}$) to extract high-level features $z_{2}$. Finally, multi-scale semantic features are integrated from \textbf{TMSN} (Trusted Multi-Scale Network) and aggregated to generate the prediction $y$.}
	\label{fig.framework}
\end{figure*}

\section{Methodology}
The overall pipeline of our framework is shown in Fig. \ref{fig.framework}, which consists of two models of transformer architectures (Low Resolution Network, High Resolution Network) and they are combined using a trusted manner. The Low Resolution Network (LRH) takes the low resolution as the input, while the High Resolution Network (HRH) aims to extract rich features from the high resolution input.
The main motivation of this framework is to imitate the actual diagnosis process of pathologists. In the diagnosis process, pathologists firstly find out the suspected region by scanning low-resolution WSIs and then check the suspected region using high-resolution patches. After that, pathologists provide the diagnosis result by jointly leveraging the characteristics of high and low-resolution images. We will explain the components in more details subsequently.

\subsection{Low Resolution Network}
LRN consists of two main sub-components: the Vision Transformer encoder and Attention Rollout.
Here, we adopt the ViT-S/32 \cite{dosovitskiy2020image} as the backbone network: $\phi_l$, which project the down-scaled WSI, $x \in \mathbb{R}^{H \times W \times C}$, into feature vector $z_1$ as:
\begin{equation}
	\phi_{l}(x)=z_{1},
\end{equation}
\noindent where  $z_{1} \in \mathbb{R}^{1 \times d}$ and $d = 384$ is the output of the penultimate layer of the Vision Transformer architecture.

\begin{algorithm}[!t]
	\caption{Non-Maximum Suppression Patch Selection}\label{alg:cap}
	\begin{algorithmic}[1]
		\Require patches set $P=\left\{P_{1}, \ldots, P_{Q}\right\}$ corresponding to the attention map, which $S_{i} \in \mathbb{R}^{H \times W \times C}$; selected patches $\mathcal{X}$; the attention value corresponding to selected patch $\mathcal{A}$
		the attention rollout map $\tilde{A}$; the number of patches to be selected $N_{h}$.
		\Ensure $\mathcal{X}$, $\mathcal{A}$
		\State $\mathcal{X} \gets \left\{\right\}$, $\mathcal{A} \gets \left\{\right\}$, $i \gets 0$
		\While{$i < N_{h}$}
		\State $m \gets argmax(\tilde{A})$ 
		% $\Statex \Comment{Finding highest attention patch.}
		\State $\mathcal{B}_{p} \leftarrow P_{m}$
		\State $\mathcal{B}_{att} \leftarrow \tilde{A}_{m}$
		\State $\mathcal{X} \gets \mathcal{X} \cap \mathcal{B}_{p}$
		\State $\mathcal{A} \gets \mathcal{A} \cap \mathcal{B}_{att}$ 
		\State $\tilde{A}_{m,tblr} \gets 0$  
		% \Statex \Comment{Setting 0 to top, bottom, left, right patches around patch $m$ in $\tilde{A}_{m}$.}
		\State $i \gets i + 1$
		\EndWhile
	\end{algorithmic}
\end{algorithm}

In the Vision Transformer, both the attention weights and the residual connections play an important role in the process of propagates information. Therefore, attention rollout (AR) \cite{abnar2020quantifying} can use additional weights to represent the residual connections in the attention graph.
Specifically, assuming that the Vision Transformer has $L$ layers, for the layer $l$ with residual connection, we compute values in layer $l$ as $V_{l}= \left( W_{att,l} + I \right) V_{l-1} $, where $W_{att, l}$ represents the attention matrix in layer $l$. Therefore, an identity matrix employed to account for residual connections. Thus, the re-normalize attention can be formulated as $A(l)= 0.5 \times (W_{att, l} + I)$. The flow in Vision Transformer can propagate from the input layer to the embedding layers. To obtain this information flow relationship, AR recursively multiply the attention matrices in all the layers:

\begin{equation}
	\tilde{A}\left(l\right)= \begin{cases}A\left(l\right) \tilde{A}\left(l-1\right) & \text { if } l>0 \\ A\left(l\right) & \text { if } l=0\end{cases}
\end{equation}
\noindent where $\tilde{A}$ is attention rollout. Each value of attention rollout at top layer indicates the importance of the region at the corresponding position. It is worth noting that we use the attention value of the last layer for subsequent operations.

\subsection{High Resolution Network}
In the High Resolution Network (HRN) branch, we used AR to compute the attention maps.
However, AR may leads excessive concentration in a certain region when selects them based on sorted attention value. Thus, the model may neglect other discriminative patches. For this reason, we use Non-Maximum Suppression Patch Selection (NMS-PS) to select patches.

As shown in Algorithm 1, $N_{h}$ can be selected under the constraints of the GPU capacity, and $\tilde{A}_{m,tblr}$ means setting 0 to top, bottom, left, right patches around patch $m$ in $\tilde{A}_{m}$. NMS-PS can obtain the sparse task-related high-resolution patches bag with $\mathcal{X}$ and corresponding attention vector $\mathcal{A}$. 

Then, we can extract high resolution representation with the following two modules. The encoder structure $\phi_{h}$ also uses ViT-S/32, which converting the patches in the bag into feature vectors $\mathcal{I} = \left\{h_{1}, \ldots, h_{N_{h}}\right\}, h_{i}\in \mathbb{R}^{1 \times d}$. LRN and HRN do not share the parameter as traditional methods provide low robustness to scale changes \cite{singh2018analysis}. Then we employ the attention mechanism \cite{ilse2018attention} to ensemble the feature vectors in the bag as high resolution representation $z_{2}$. Thus, HRN can be formulated as:

\begin{equation}
	\phi_{h}(\mathcal{X})=\mathcal{I},
\end{equation}

\begin{equation}
	z_{2}=\sum_{i \in \mathcal{I}} a_{i} h_{i} \mathcal{A}_{i},
\end{equation}

\begin{equation}
	a_{i}=\frac{\exp \left(\boldsymbol{w}^{\top} \tanh \left(\boldsymbol{V} h_{i} \mathcal{A}_{i}\right)\right)}{\sum_{j \in \mathcal{I}} \exp \left(\boldsymbol{w}^{\top} \tanh \left(\boldsymbol{V} h_{j} \mathcal{A}_{j}\right)\right)}, i \in \mathcal{I},
\end{equation}

\noindent where $\boldsymbol{w}$ and $\boldsymbol{V}$ are trainable parameters for attention network. The multi-scale representation of the WSI, $(z_{1}, z_{2})$, are extracted and input into the TMSN. The details of TMSN are described in detail below.

\subsection{Trusted Multi-Scale Network}

It's difficult to obtain a trusted result from the single scale for the WSI classification task. To this end, we employed trusted multi-view learning \cite{han2021trusted}, which can automatically integrate evidence from different scales to improve the robustness and reliability of the network.

We calculate the evidence based on feature vectors to obtain the probabilities and the uncertainty of $K$-classification problems for low and high resolution branch. Patches at different scales have distinct representations as the WSI is too large. Therefore, we regard high and low resolution as two views, respectively. For the $v^{th}$ view, the evidence $e^{v}$ is defined as $ e^{v} = \sigma \left( z_{v} \boldsymbol{w}^{T}_{v} + \boldsymbol{b}_{v}\right) $, where $w_{v} \in \mathbb{R}^{K \times d}$ and $b_{v} \in \mathbb{R}^{K}$ are trainable parameters, $\sigma$ is the Softplus function \cite{nwankpa2018activation}, which ensures that the evidence value is non-negative.
For the $v^{th}$ view, the Dirichlet distribution is used to model the class distribution according to the evidence $e^{v}=\left(e^{v}_{1}, ... , e^{v}_{K}\right)$. the Dirichlet distribution parameters are defined as: $\alpha^{v} = \left(\alpha^{v}_{1}, ... , \alpha^{v}_{K}\right)$, where $\alpha^{v}_{k} = e^{v}_{k}+1$. Then, the class probability $b^{v}_{k}$ and the uncertainty $u^{v}$ are formulated as:

\begin{equation}
	b_{k}^{v}=\frac{e_{k}^{v}}{S^{v}}=\frac{\alpha_{k}^{v}-1}{S^{v}} \quad \text { and } \quad u^{v}=\frac{K}{S^{v}},
	\label{form.bu}
\end{equation}

\noindent where $ S^{v}=\sum_{i=1}^{K} \alpha_{i}^{v} $ is the Dirichlet strength, 
and the sum of class probability $b^{v}_{k}$ and the uncertainty $u^{v}$ is one. 
We formulate mass $\mathcal{M}^{v}$ as $\left\{ \{b_{k}^{v} \}_{k=1}^{K}, u^{v}\right\}$.
The uncertainty is lower when the Dirichlet distribution parameter is sharp distribution, for example, $ \left< 1, 0, 0 \right>$.
The uncertainty will be more higher when the parameter is flat distribution, for example, $\left< 1, 1, 1 \right>$.
The Dempster–Shafer theory is used to ensemble the results from different sources based on the class probability and uncertainty. The ensemble rule of two views 
$\mathcal{M} = \left\{ \{b_{k}\}_{k=1}^{K}, u\right\}$ 
are defined as:

\begin{small}
	\begin{equation}
		b_{k}=\frac{1}{1-C}\left(b_{k}^{1} b_{k}^{2}+b_{k}^{1} u^{2}+b_{k}^{2} u^{1}\right), u=\frac{1}{1-C} u^{1} u^{2},
	\end{equation}
\end{small}

\noindent where $C=\sum_{i \neq j} b_{i}^{1} b_{j}^{2}$ is the amount of conflict between two source sets. Then, we can calculate the ensemble Dirichlet distribution parameters $\alpha$ according to Eq. \ref{form.bu}. Intuitively, a source view with low uncertainty will contribute more to the ensemble result. 

Benefiting from the aforementioned rules, we can obtain the estimated multi-view joint distribution to generate the final uncertainty and the ensemble probability of each class.

\subsection{Aggregation and Loss Functions}
Based on the semantic features $\left\{ z^{1}, z^{2} \right\}$, the Dirichlet distribution parameters and corresponding masses $\left\{ \mathcal{M}, \mathcal{M}_{1}, \mathcal{M}_{2} \right\}$ obtained from previous modules, our model be optimized with trusted multi-scale loss.

The classification task usually employs the cross-entropy loss function: $\mathcal{L}_{c e}=-\sum_{i=1}^{K} y_{i} \log \left(p_{i}\right)$, where $p_{i}$ is the predicted probability for class $i$. Based on the evidence $e^{th}$ obtained from $v^{th}$, we can obtain Dirichlet distribution parameter $\alpha^{th}$ and form the opinions $D ( \mathbf{p} | \boldsymbol{\alpha}^{th} )$, where $\mathbf{p}$ is the class probabilities. With modification on $\mathcal{L}_{ce}$, we have adjusted cross-entropy loss as follows:

\begin{equation}
	% \mathcal{L}_{a c e}\left(\boldsymbol{\alpha}_{i}\right)=\int\left[\sum_{j=1}^{K}-y_{i j} \log \left(p_{i j}\right)\right] \frac{1}{B\left(\boldsymbol{\alpha}_{i}\right)} \prod_{j=1}^{K} p_{i j}^{\alpha_{i j}-1} d \mathbf{p}_{i}=\sum_{j=1}^{K} y_{i j}\left(\psi\left(S_{i}\right)-\psi\left(\alpha_{i j}\right)\right)
	\mathcal{L}_{a c e}\left(\boldsymbol{\alpha}^{th}\right)=\sum_{i=1}^{K} y_{i}\left(\psi\left(S^{th}_{i}\right)-\psi\left(\alpha^{th}_{i}\right)\right),
\end{equation}

\noindent where $\psi$ is the diagram function. This loss item encourages that the correct class will be generated more evidences than incorrect classes. In the meantime, the Dirichlet distribution parameter $\alpha^{th} $ is closer to the sharp distribution, which decrease incorrect classes evidences to 0. To this end, KL divergence term was employed:

\begin{algorithm}[!t]
	\caption{The training pipeline of our proposed method.}\label{alg:cap}
	\begin{algorithmic}[1]
		\Require The WSI dataset $\left\{\mathbb{P}_{n}, \left(\mathbb{X}_{n}, \mathbb{Y}_{n}\right)\right\}_{n=1}^{N}$; 
		\Statex \quad \ \ The number of patches $N_{h}$ to be selected; 
		\Statex \quad \ \ Loss factors $ \lambda $.
		\Ensure The trained network.
		\While{model reaches convergence}
		\For{each $P$, $x$, $y$ in all data sets}
		
		\State Extract low-resolution features $z_{1}$ using $\phi_{l}$;
		
		\State Obtain attention map $\tilde{A}$ by inference of AR;
		
		\State Generate high-resolution patches bag $\mathcal{F}$ and 
		\Statex \qquad \quad \ \ \ corresponding attention values set $\mathcal{A}$ using 
		\Statex \qquad \quad \ \ \ NMS-PS.
		
		\State Extract high-resolution features bag $\mathcal{I}$ by 
		\Statex \qquad \quad \ \ \ inference of $\phi_{h}$;
		
		\State Aggregate features bag $\mathcal{I}$ to high-
		\Statex \qquad \quad \ \ \ resolution features $z_{2}$ using multi-instance 
		\Statex \qquad \quad \ \ \ attention mechanism;
		
		\State Ensemble multi-scale features by inference 
		\Statex \qquad \quad \ \ \ of TMSN;
		
		\State Minimize $\mathcal{L}$ in Eq. \ref{eql.loss}.
		
		\EndFor
		
		\State update $\lambda$.
		
		\EndWhile
	\end{algorithmic}
	\label{algo.all}
\end{algorithm}

%\begin{equation}
% \begin{aligned}
% &KL\left[D\left(\mathbf{p}_{i} \mid \tilde{\boldsymbol{\alpha}}_{i}\right) \| D\left(\mathbf{p}_{i} \mid \mathbf{1}\right)\right]\\
% &=\log \left(\frac{\Gamma\left(\sum_{k=1}^{K} \tilde{\alpha}_{i k}\right)}{\Gamma(K) \prod_{k=1}^{K} \Gamma\left(\tilde{\alpha}_{i k}\right)}\right)+\\
% &\sum_{k=1}^{K}\left(\tilde{\alpha}_{i k}-1\right)\left[\psi\left(\tilde{\alpha}_{i k}\right)-\psi\left(\sum_{j=1}^{K} \tilde{\alpha}_{i j}\right)\right]
% \end{aligned}
% \end{equation}

%\begin{equation}
%\mathcal{L}\left(\boldsymbol{\alpha}_{i}\right)=\mathcal{L}_{a c e}\left(\boldsymbol{\alpha}_{i}\right)+\lambda_{t} K L\left[D\left(\mathbf{p}_{i} \mid \tilde{\boldsymbol{\alpha}}_{\boldsymbol{i}}\right) \| D\left(\mathbf{p}_{i} \mid \mathbf{1}\right)\right]
%\end{equation}

% \begin{small}
\begin{equation}
	\mathcal{L}_{v} (\boldsymbol{\alpha}^{th} )=\mathcal{L}_{a c e} (\tilde{\boldsymbol{\alpha}}^{th} )+\lambda K L [ D ( \mathbf{p} | \boldsymbol{\alpha}^{th} ) \| D ( \mathbf{p} | \mathbf{1} ) ],
\end{equation}
% \end{small}

\noindent where $\lambda > 0$ is the balance coefficient to prevent the network paying too much attention to the KL divergence, and $\tilde{\boldsymbol{\alpha}}^{th} = y + (1-y) \odot \boldsymbol{\alpha}^{th}$ is the modified parameter of the Dirichlet distribution which can avoid punishing the evidence of the ground truth class to 0. In order to ensure that all views are optimized simultaneously, the total loss is formulated as:

%\begin{equation}
%\mathcal{L}_{\text {da}}=\sum_{i=1}^{N}\left[\mathcal{L}\left(\boldsymbol{\alpha}_{i}\right)+\sum_{v=1}^{V} \mathcal{L}\left(\boldsymbol{\alpha}_{i}^{v}\right)\right]
%\end{equation}

\begin{equation}
	\mathcal{L}=
	\mathcal{L}_{v}\left(\boldsymbol{\alpha}\right)+ 
	\mathcal{L}_{v}\left(\boldsymbol{\alpha}^{1}\right)+
	\mathcal{L}_{v}\left(\boldsymbol{\alpha}^{2}\right),
	\label{eql.loss}
\end{equation}

Through optimizing trusted multi-scale WSI classification with Eq. \ref{eql.loss}, our method can obtain the exceptional performance with selective high resolution information. 
The optimization process is summarized in Algorithm \ref{algo.all}.

\section{Experiments}
We present our experiments results in this section. We first outline the evaluation settings regarding the WSI classification task. Then we conduct the experiments on two WSI datasets: LKS \footnote{https://github.com/cradleai/LKS-Dataset} and our Fibroma WSI dataset.
Moreover, as our framework consists of several improvements, we attempt to tease apart the factors that affect the WSI classification performance, using an extensive ablation studies.

\subsection{Experiment setup}
Four metrics are employed to evaluate the performance, including accuracy (ACC), macro-F1 score (F1), Kappa score (Kappa) and Matthews correlation coefficient (MCC) \cite{boughorbel2017optimal}. Specifically, the macro-F1 score can balance the precision and recall and it is suitable for evaluation imbalanced datasets, while the MCC score can be used to describe the correlation between the predictions and the labels.

The dataset has been divided into training set and test set, according to the ratio of 7:3. Moreover, we split the training set mentioned above into 80\% training data and 20\% validation data. To pre-process the WSI datasets, each WSI is scaled to $384 \times 384$ as the LRN input, and HRN input size is consistent with LRN. As aforementioned, we utilized the ViT-S/32 as the feature extractor, which converts the image into 384-dimensional feature vector. Meanwhile, AR can obtain the attention map with size of $12 \times 12$, and $Q$ in NMS-PS is equal to 144. The OTSU algorithm \cite{otsu1979threshold} is deployed to generate the foreground mask to obtain the attention map generated by AR. For the practical implementation, we employed PyTorch framework \cite{paszke2019pytorch} and we trained our models until convergence on a single NVIDIA 2080Ti GPU. The network optimized by SGD with the batch size of 4. The learning rate is initialized as 1e-3 and the $\lambda$ in HRN set as: $min(1, now\_epoch\_number / 10)$.

\begin{table}[!t]
	\centering
	\begin{tabular}{cccccc}
		\toprule[0.5pt]
		Set & Neg & AMA & SMA-V & SMA-T & Total\\ \midrule[0.5pt]
		Train & 239 & 106 & 107 & 27 & 479 \\
		Test & 103 & 45 & 46 & 11 & 205 \\ \bottomrule[0.5pt]
	\end{tabular}
	\caption{The distribution of classes in the LKS dataset, which consists of four sub-classes of antibody, Negative (Neg), Anti-Mitochondrial Antibodies (AMA), Vessel-Type Anti-Smooth Muscle Antibodies (SMA-V) and Tubule-Type Anti-Smooth Muscle Antibodies (SMA-T).}
	\label{label.lks}
\end{table}

\begin{table}[!t]
	\centering
	\begin{tabular}{cccc}
		\toprule[0.5pt]
		Set & DF & NF & Total\\ \midrule[0.5pt]
		Train & 347 & 192 & 539  \\
		Test & 150 & 82 & 232 \\ \bottomrule[0.5pt]
	\end{tabular}
	\caption{The distribution of classes in the Fibroma dataset, which includes 497 Dermatofibroma (DF) WSIs and 274 Neurofibromatosis (NF) WSIs.}
	\label{label.fib}
\end{table}

\subsection{Baseline Methods}
We introduce the baseline details for comparison below.

\noindent \textbf{WSI-Level.}
The Image-Level method conduct the classification using single-scale images. Specifically, we employ WSI thumbnails to train a ViT-S/32 \cite{dosovitskiy2020image} model for fair comparison.

\noindent \textbf{Patch-Level.}
The patch level-based method executes the classification with patch-level images. The network structure is basically as same as the HRN of the proposed method. However, we replace Attention Rollout and NMS-PS with random cropping. Therefore, only the selected $K$ patches at high resolution are used to classify WSI.

\noindent \textbf{Multi-Scale.}
The Multi-Level method uses both the image-level features and patch-level features for classification. The Multi-Level method design is parallelism with the Patch-level network. However, we concatenate the image-level features and the patch-level features together through the multi-instance attention module. Therefore, both high resolution feature and low resolution feature are used to classify WSI.

\begin{figure}[!t]
	\centering
	\includegraphics[scale=0.37]{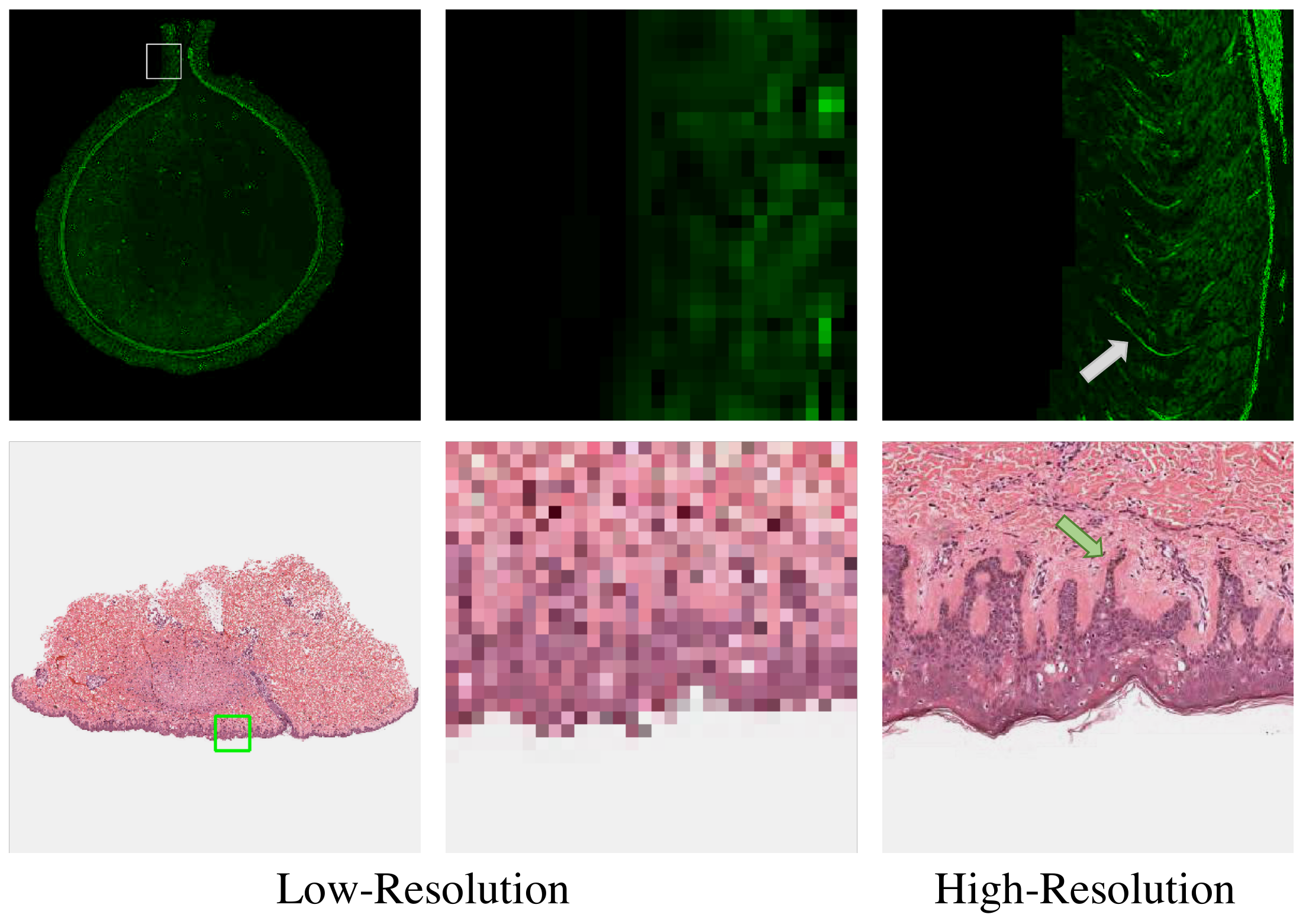}
	\caption{
		Examples of WSI patches sampled by the HRN. The second column of images extracted from the color boxes in the first column. The white and green arrows indicate the discriminative representations of certain classes. 
	}
	\label{fig.vis3}
\end{figure}

\begin{table*}[!t]
	\centering
	\begin{tabular}{c|cccc|cccc}
		\toprule[0.5pt]
		& \multicolumn{4}{c|}{LKS} & \multicolumn{4}{c}{Fibroma} \\ \midrule[0.5pt]
		\midrule[0.5pt]
		Method & ACC(\%) & F1(\%) & Kappa(\%) & MCC(\%) & ACC(\%) & F1(\%) & Kappa(\%) & MCC(\%) \\ 
		\midrule[0.5pt]
		\midrule[0.5pt]
		Image-Level & 84.88 & 80.97 & 77.19 & 77.53 & 83.19 & 81.85 & 63.72 & 63.79 \\ 
		Patch-Level & 79.02 & 64.78 & 68.33 & 69.77 & 76.72 & 76.33 & 53.91 & 56.99 \\
		Multi-Scale & 88.78 & 84.07 & 82.58 & 82.60 & 87.07 & 85.51 & 71.07 & 71.28 \\
		\midrule[0.5pt]
		PTree-Net & 89.76 & 83.54 & 84.18 & 84.22 & 88.36 & 86.84 & 73.74 & 74.16 \\
		SOS & 90.73 & 87.47 & 85.85 & 85.94 & 88.79 & 87.74 & 75.48 & 75.48 \\
		Our & \textbf{92.68} & \textbf{87.79} & \textbf{88.79} & \textbf{88.86} & \textbf{92.24} & \textbf{91.36} & \textbf{82.74} & \textbf{82.88} \\
		\bottomrule[0.5pt]
	\end{tabular}
	\caption{The comparison between the proposed methods and state-of-the-art methods on two datasets. Image-Level, Patch-Level, Multi-Scale, SOS and our proposed method were compared, and our purposed trusted multi-scale model showed the highest accuracy.}
	\label{label.result}
\end{table*}

\begin{table*}[!t]
	\centering
		\begin{tabular}{cccc|cccc|cccc}
			\toprule[0.5pt]
			\multicolumn{4}{l|}{} & \multicolumn{4}{c|}{LKS} & \multicolumn{4}{c}{Fibroma} \\ \midrule[0.5pt]
			\midrule[0.5pt]
			NMS-PS & Fusion & Attention & K & ACC(\%) & F1(\%) & Kappa(\%) & MCC(\%) & ACC(\%) & F1(\%) & Kappa(\%) & MCC(\%) \\ 
			\midrule[0.5pt]
			\midrule[0.5pt]
			\checkmark & TMSN & AR & 4 & 90.24 & 86.46 & 85.03 & 85.06 & 89.66 & 88.62 & 77.24 & 77.25 \\
			\checkmark & TMSN & AR & 6 & 91.71 & 87.69 & 87.36 & 87.44 & 90.52 & 89.63 & 79.25 & 79.25 \\
			\checkmark & TMSN & AR & 8 & 92.20 & \textbf{87.90} & 88.05 & 88.10 & 91.38 & 90.40 & 80.82 & 80.95 \\ 
			\checkmark & TMSN & AR & 10 & \textbf{92.68} & 87.79 & \textbf{88.79} & \textbf{88.86} & \textbf{92.24} & \textbf{91.36} & \textbf{82.74} & \textbf{82.88} \\
			\midrule[0.5pt]
			\checkmark & TMSN & F-AR & 4 & 89.76 & 84.98 & 84.07 & 84.10 & 88.79 & 87.36 & 74.78 & 75.14 \\
			\checkmark & TMSN & HA & 4 & 88.29 & 82.70 & 82.21 & 82.32 & 87.07 & 85.60 & 71.23 & 71.35 \\
			\midrule[0.5pt]
			\checkmark & AVG & AR & 10 & 90.24 & 84.76 & 85.10 & 85.27 & 90.09 & 89.06 & 78.13 & 78.16 \\
			& TMSN & AR & 10 & 91.22 & 87.23 & 86.39 & 86.43 & 90.52 & 89.44 & 78.90 & 79.03\\
			
			\bottomrule[0.5pt]
	\end{tabular}
	\caption{Ablation study of our proposed methods. AR/F-AR denote attention values from HRN can/can't be optimized by LRN.}
	\label{label.ablation}
\end{table*}

\noindent \textbf{Selective Objective Switch (SOS).}
The SOS \cite{maksoud2020sos} method uses both the image-level features and patch-level features for WSI classification, which is state-of-the-art for this task. For quantitative comparison, we reproduced the SOS network \cite{maksoud2020sos}.

\noindent \textbf{PTree-Net.}
The PTree-Net \cite{chen2021diagnose} method uses both the image-level features and patch-level features for WSI classification. Compared with other multi-scale two level method (Level 1: low, Level 2: high), PTree-Net extracts the patches at three levels (Level 1: low, Level 2: high, Level 3: higher). It is worth noting that PTree-Net uses ResNet18 as the feature extraction network instead of Vision Transformer, because it's attention module, class activation maps (CAM) \cite{zhou2016learning}, does not support Visual Transformer presently.

\subsection{Experimental Results Using the LKS datasets}
We first present our results on LKS dataset. We introduce the dataset and discuss our results on classification.

\noindent \textbf{Datasets.}
LKS is a public dataset purposed for antibodies detection in autoimmune diseases. This dataset consists of four sub-classes of antibody, Negative (Neg), Anti-Mitochondrial Antibodies (AMA), Vessel-Type Anti-Smooth Muscle Antibodies (SMA-V) and Tubule-Type Anti-Smooth Muscle Antibodies (SMA-T), with in a total of 684 diagnostic digital slides. The category distribution is shown in Table \ref{label.lks}.

\noindent \textbf{Results.} 
The number of patches to be selected $N_{h}$ is set to 10 in Patch-Level, for our proposed method and SOS method. The classification results are summarized in Table \ref{label.result}. The results show that: though both better than Multi-Scale method, our method achieves better feature aggregation performance than SOS, with about 2.92\% improvements in MCC on the multi-scale. At low resolution, some discriminative features are difficult to identify, as showed in Fig. \ref{fig.vis3}. The faultiness can't identify these discriminative features may explain why the F1 and Kappa scores of these classes are using Image-Level method obvious lower than using Multi-Scale, SOS, PTree-Net and our methods. Random cropping will generate a large number of useless patches. Thus, Patch-Level method is difficult to obtain the discriminating patch to classification, with about 12.8\% drop in accuracy.

\begin{figure*}[!t]
	\centering
	\includegraphics[scale=0.56]{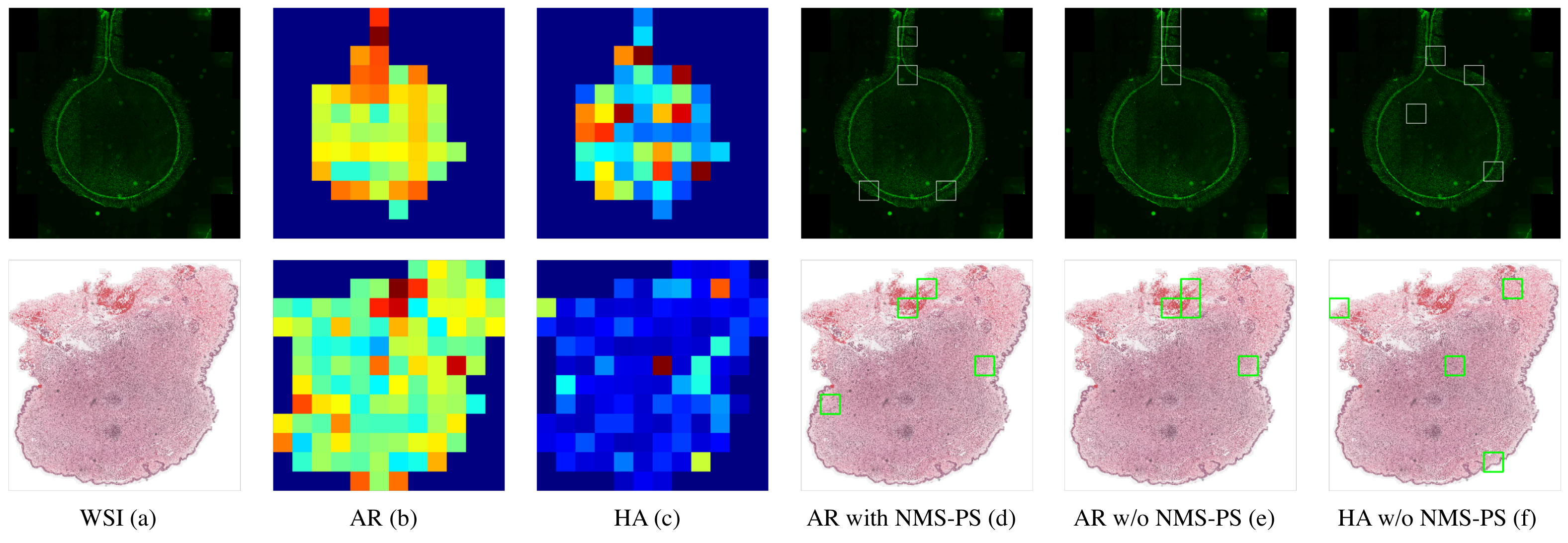}
	\caption{
		Visualization of attention weights and patch selection under different methods in high-resolution. The top row represents the results of the LKS dataset, and the bottom row represents the results of the Fibroma dataset. Note that the attention map of AR and HA all based generated using our purposed model. The white and green boxes represent the discriminative patches selected by different algorithms with different datasets, respectively.
	}
	\label{fig.vis2}
\end{figure*}

\subsection{WSI of Fibroma Classification}

\noindent \textbf{Datasets.}
To further demonstrate the effective of proposed method, we also conduct extensive experiments on the newly collected dataset. Specifically, these data were collected from Shanghai Skin Disease Hospital and scanned using multiple image scanners. All tissues and data were retrieved under the permission of the institutional research ethics committee of the institution. The datasets consists of 771 hematoxylin-eosin (HE) stained WSI images, including 497 Dermatofibroma (DF) WSIs and 274 Neurofibromatosis (NF) WSIs. The category distribution is shown in Table \ref{label.fib}. Each WSI is labeled by an experienced skin pathologist.

\noindent \textbf{Results.}
As shown in Table \ref{label.result}, our model achieves the best performance on four evaluation metrics, including the accuracy of 92.24\%, macro-F1 of 91.36\%, kappa
of 82.47\% and MCC of 82.88\%. Compared with the Image-Level single-scale method, the multi-scale information and well-designed framework bring a 9\% accuracy and 19\% kappa improvement to our model.
Compared with the PTree-Net, SOS and our methods bring 2.92\% and 0.97\% improvement of accuracy, respectively. We suspect the main reason for the poor results of PTreeNet is the patches selected at WSI Level 2 less than SOS and our method, and the patches selected at Level 2 is more important than Level 3. The results suggest similar conclusions as LKS dataset. Following these comparisons, our proposed method explores multi-scale information more robust and integrates the multi-scale features more trusted, which achieves the outperforming performance in the WSI classification task.

\subsection{Ablation Study}
We now represent the contributions of our model via ablation studies of the two major components of our model: the mechanism of patch selection and trusted multi-scale classification framework.

\noindent \textbf{Effects of Patch Selection.}
In order to verify the effectiveness of our method, we demonstrate the heatmap of AR and HA (hard attention mechanism) \cite{maksoud2020sos} at two datasets, respectively.
As shown in Fig. \ref{fig.vis2} (b, c), the attention heatmap generated by AR is focuses on some certain patches and the attention value decreases to the surroundings, which keep consistent with the characteristics of human vision. 
Thus, in Fig. \ref{fig.vis2} (e, f), compared with HA, the AR selected patches based on attention values are more concentrated.
The patches selected by NMS-PS exist in more attention centers compared with the highest attention value based method (Fig. \ref{fig.vis2} (d, e)). However, adjacent patches in WSI often have similar representations, therefore, NMS-PS can obtain more high resolution discriminative patches. We also conducted experiments to verify the effectiveness of the proposed method in Table \ref{label.ablation}. 
The comparison among attention methods indicates that F-AR brings 1.7\% accuracy increase over HA on Fibroma dataset, and the knowledge of LRN to update AR improves F1 with 1.26\%.

\noindent \textbf{Effects of Trusted Multi-Scale Classification.}
To further verify the effectiveness, we also compared our TMSN with previous work. We replace TMSN with decision-level averaging fusion method (AVG) \cite{simonyan2014two} to ensemble multi-scale classification results, which proves that the usage of TMSN framework contributes 3.6\% improvement to MCC on the LKS dataset. Compared with AVG, TMSN makes the final decision based on the uncertainty when the discriminative patches in the high resolution are not selected, as shown in Fig. \ref{fig.aim}.

\section{Conclusion}
In this paper, we proposed a trusted multi-scale classification framework to exploit the multi-scale feature of WSI, leveraging the Vision Transformer as the backbone. Our framework can jointly classification modeling and estimating the uncertainty of each magnification of a microscope and integrate the evidence from different magnification. Moreover, a novel patch selection method (NMS-PS) is proposed to exploit discriminative patches from WSIs and reduce the requirement for computation resources. With NMS-PS selected patches from the attention map, we employ the multi-instance attention mechanisms to ensemble the high-level features. Finally, TMSN is used to fusion multi-scale decision based on uncertainty. Extensive experiments show that the trusted framework can significantly improve the WSI classification performance compared with the state-of-the-art methods.

% \begin{acks}
% \end{acks}

\bibliographystyle{ACM-Reference-Format}
\bibliography{sample-base}


\end{document}